\documentclass[twocolumn]{ceurart}
\begin{document}

\copyrightyear{2023}
\copyrightclause{Copyright for this paper by its authors.
  Use permitted under Creative Commons License Attribution 4.0
  International (CC BY 4.0).}

\conference{Proceedings of the Sixth Workshop on Automated Semantic Analysis of Information in Legal Text (ASAIL 2023), June 23, 2023, Braga, Portugal.}

\title{Contrast Is All You Need}

\author[1, 2]{Burak Kilic}[
    email=b.kilic@uu.nl,
    url=https://www.shareforcelegal.com,
]
\cormark[1]
\address[1]{Department of Information and Computing Sciences, Utrecht University, Utrecht, The Netherlands}
\address[2]{Shareforce B.V., Rotterdam, The Netherlands}
\address[3]{Tilburg Institute for Law, Technology, and Society, Tilburg University, Tilburg, The Netherlands}

\author[1, 3]{Floris Bex}[
    email=f.j.bex@uu.nl,
    url=https://www.uu.nl/staff/FJBex,
]
\fnmark[1]

\author[1]{Albert Gatt}
[
    email=a.gatt@uu.nl,
    url=https://albertgatt.github.io/,
]

\fnmark[1]

\cortext[1]{Corresponding author.}
\fntext[1]{These authors contributed equally.}

\begin{abstract}
 In this study, we analyze data-scarce classification scenarios, where available labeled legal data is small and imbalanced, potentially hurting the quality of the results. We focused on two finetuning objectives; SetFit (Sentence Transformer Finetuning), a contrastive learning setup, and a vanilla finetuning setup on a legal provision classification task. Additionally, we compare the features that are extracted with LIME (Local Interpretable Model-agnostic Explanations) to see which particular features contributed to  the model's classification decisions. The results show that a contrastive setup with SetFit performed better than vanilla finetuning while using a fraction of the training samples. LIME results show that the contrastive learning approach helps boost both positive and negative features which are legally informative and contribute to the classification results. Thus a model finetuned with a contrastive objective seems to base its decisions more confidently on legally informative features.

\end{abstract}


\maketitle

\section{Introduction}

 The scarcity of publicly available, high quality legal data is causing a bottleneck in legal text classification research~\cite{DBLP:journals/corr/abs-1911-00473}. While there are a few publicly available datasets, such as CUAD~\cite{DBLP:journals/corr/abs-2103-06268}, and LEDGAR~\cite{Tuggener2020LEDGARAL}, these datasets are unbalanced. They may provide good baselines to start with; however, the scarcity of samples for specific classes means that there is no guarantee of robust performance once models are adapted to downstream classification tasks.

Few-shot learning methods have proven to be an attractive solution for classification tasks with small datasets where data annotation is also time-consuming, inefficient and expensive. These methods are designed to work with a small number of labeled training samples and typically require adapting a pretrained language model to a specific downstream task. 

In this paper~\footnote{While our paper shares a similar title with "Attention is all you need"~\cite{DBLP:journals/corr/VaswaniSPUJGKP17}, we focus on a different topic.}, we focus on three major aims. First, we finetune the LegalBERT\cite{chalkidis-etal-2020-legal} model on the publicly available LEDGAR provision classification dataset. We compare the success of a contrastive learning objective and a more standard objective to finetune the pretrained model.

Secondly, we finetune the same baseline model with these two finetuning objectives with the balanced dataset created from LEDGAR. Lastly, to analyze the trustworthiness and explain individual predictions, we extract the tokens from the model as features by using LIME~\cite{lime} to compare which features had more positive or negative impacts. 

\section{Related Work}

The legal text classification has been tackled with various BERT techniques to adopt domain-specific legal corpora~\cite{limsopatham-2021-effectively}~\cite{10.1145/3086512.3086515}. While these studies often report state-of-the-art results with BERT-based models, they do not address the issue of data scarcity for specific applications. 

There have been several pieces of research on efficient finetuning setups that can potentially address this necessity, such as parameter efficient finetuning (PEFT), pattern exploiting training (PET), and SetFit (Sentence Transformer Finetuning)~\cite{tunstall2022efficient}, an efficient and prompt-free framework for few-shot finetuning of Sentence Transformers (ST). SetFit works by first finetuning a pretrained ST on a small number of text pairs, in a contrastive Siamese manner. Also, SetFit requires no prompts or verbalizers, unlike PEFT and PET. This makes SetFit simpler and faster. We explain how SetFit works in more depth in the following section.

\subsection{SetFit: Sentence Transformer Finetuning}

SetFit is a prompt free framework for few-shot finetuning of ST, addressing labeled data scarcity by introducing contrastive learning methods to generate positive and negative pairs from the existing dataset to increase the number of samples. 

There are two main steps involved in SetFit, from training to inferencing. First, a contrastive objective is used to finetune the ST, and then the classification head is trained with the encoded input texts.

At the inference stage, the finetuned ST also encodes the unseen inputs and produces the embeddings accordingly. Then the classifier head gives the prediction results based on the newly generated embeddings.

\paragraph{ST finetuning} To better handle the limited amount of labeled training data in few-shot scenarios, contrastive training approach is used. Formally, we assume a small set of K-labeled samples $D = {(x_i, y_i)}$, where $x_i$ and $y_i$ are sentences and their class labels, respectively. For each class label $c \in C$, $R$ positive triplets are generated: $T_{p}^{c}={(xi , xj , 1)}$, where $x_i$ and $x_j$ are pairs of randomly chosen sentences from the same class $c$, such that $y_i = y_j = c$. Similarly, a set of $R$ negative triplets are also generated: $T_{n}^{c}={(xi, xj , 0)}$, where $x_i$ are sentences from class $c$ and $x_j$ are randomly chosen sentences from different classes such that $y_i = c$ and $y_j \neq c$. Finally, the contrastive finetuning data set $T$ is produced by concatenating the positive and negative triplets across all classes where $|C|$ is the number of class labels, $|T| = 2R|C|$ is the number of pairs in $T$ and $R$ is a hyperparameter. SetFit will generate positive and negative samples randomly from the training set, unless they are explicitly given~\cite{tunstall2022efficient}.

This contrastive finetuning approach enlarges the size of training data. Assuming that a small number ($K$) of labeled samples is given for a binary classification task, the potential size of the ST finetuning set $T$ is derived from the number of unique sentence pairs that can be generated, namely $K(K - 1)/2$, which is significantly larger than just $K$.

\paragraph{Classification head training} In this second step, the fine-tuned ST encodes the original labeled training data $\{x_i\}$, yielding a single sentence embedding per training sample: $Emb(x_i) = ST(x_i)$ where $ST()$ is the function representing the fine-tuned ST. The embeddings, along with their class labels, constitute the training set for the classification head $TCH = {(Emb(x_i), y_i)}$ where $|TCH| = |D|$. A logistic regression model is used as the text classification head throughout this work.

\paragraph{Inference} At inference time, the fine-tuned ST encodes an unseen input sentence ($x_i$) and produces a sentence embedding. Next, the classification head that was trained in the training step, produces the class prediction of the input sentence based on its sentence embedding. Formally this is
predicted label $i = CH(ST(xi))$, where $CH$ represents the classification head prediction function.

\section{Data}
\label{sec:data}
We present experimental results both on the original LEDGAR dataset, and on a balanced version.
We describe the original dataset first, then we give a brief description of how the dataset was further balanced for the presented experiments.

\subsection{Data source}
As a main corpus, we used the publicly available LEDGAR\footnote{https://autonlp.ai/datasets/ledgar} provision classification dataset, consisting of 60,000 training samples in total, with 100 different provision labels.

We did not apply any additional preprocessing or data modification techniques to keep the data as it is to make the experiments reproducible.

To create a dedicated test dataset for the unbalanced data scenario, we randomly selected 25 samples per label from the corpus, in total approximately 2,500 samples. The rest of the 57,500 samples are used to generate the train/dev sets.

The training sets are created by selecting 4, 8, 12, and 16 samples per label for SetFit, and 50, 100, 150, and 200 for the vanilla finetuning setup. Therefore the maximum number of samples is calculated as: maximum number of samples per label multiplied by the number of total labels, as can be seen in Figure~\ref{fig:acc_org}. In practice, in the case of the vanilla finetuning setup, we end up with fewer training samples than this total. This is because some labels are extremely sparse, and there are fewer total samples than the stipulated maximum per label.

\subsection{Crawling and balancing}
The original LEDGAR dataset is imbalanced. The smallest label consists of only 23 samples, and the largest has 3167 samples in the original training dataset. Therefore, to create a new balanced dataset, we selected the most frequent 32 labels.

For labels with more than 1000 samples, we downsampled to 1000 samples per label. For labels with fewer than 1000 samples, we upsampled by crawling and retrieving additional data from LawInsider,\footnote{https://www.lawinsider.com/} removing any duplicates. As a result, a new dataset has been created that consists of 32 classes, with each having 1000 provisions.

Additionally, we also created a dedicated test dataset for the balanced data scenario, and selected 25 samples per label randomly for the 32 labels, for a total of 800 samples. The remaining 31,200 samples are used for training with a random 80/20 train/dev split.

For finetuning with the balanced dataset, we again train with varying sizes of training data, using 4, 8, 12, and 16 samples per label for SetFit, and 50, 100, 150, and 200 for the vanilla finetuning setup, as can be seen in Figure~\ref{fig:acc_balanced}. 

Note that, unlike the case of the unbalanced data, the total sizes for the vanilla finetuning setup in the balanced case correspond to the totals obtained by multiplying the maximum sample size with the number of labels.

\section{Experiments}

\subsection{Models}

It has been shown that models which have been pretrained on domain-specific legal data outperform general-purpose models~\cite{chalkidis-etal-2020-legal}. Therefore, throughout this paper, the baseline we use is a finetuned LegalBERT using {\tt legal-bert-base-uncased}.\footnote{https://huggingface.co/nlpaueb/legal-bert-base-uncased} We compare this standard, or "vanilla" finetuned baseline to a model finetuned with the contrastive objective used in SetFit.

\subsection{Experimental Setup}

The finetuning setup is the most crucial stage of the experimenting setup. Therefore, we kept the common hyperparameters of SetFit and vanilla setups the same. The rest of the parameters were kept as their default values, provided by the respective implementations. The important hyperparameter for SetFit finetuning is the $R$ parameter, which defines the number of positive and negative pairs to be generated from the given training set. We kept this parameter as its default value, 20 across all the experiments. For both models, we used 1 epoch for the finetuning. Table~\ref{tab:hyperparams} gives detailed common hyperparameters of finetuning setups for both SetFit Trainer\footnote{https://github.com/huggingface/setfit} and Vanilla Trainer.\footnote{https://huggingface.co/docs/transformers/main\_classes/trainer}

\begin{table}
  \centering
  \caption{Common hyperparameters for SetFit and Vanilla Trainer} 
  \label{tab:hyperparams}
  \begin{tabular}{cc}
    \toprule
     Hyperparameter & Value \\
    \midrule
    Learning Rate & 2e-5 \\
    Warmup Ratio & 0.1\\
    Seed & 42\\
    Batch Size & 8 \\
    Epoch & 1 \\
    Metric & accuracy\\
  \bottomrule
\end{tabular}
\end{table}

\section{Results}

\subsection{F1-score comparisons: Original dataset}
In Table \ref{tab:f1-score}, we compare the F1-scores for different experiments, with the test set described in Section \ref{sec:data}. The original LEDGAR dataset is used in this experiments, with an 80/20 train/dev split. 

\begin{table}
    \centering
    \small
    \caption{F1-score comparison between SetFit and Vanilla finetuning, original LEDGAR dataset}
    \label{tab:f1-score}
    \begin{tabular}{ccccc}
        \toprule
        Models & Samples & Micro-F1 & Macro-F1 & Weighted-F1 \\ \hline
        \midrule
        Vanilla & 4933 & 0.5805 & 0.5151 & 0.5273 \\ 
        SetFit & \textbf{400} & 0.6565 & 0.6348 & \textbf{0.6423} \\ \hline
        Vanilla & 9756 & 0.6734 & 0.6180 & 0.6317 \\ 
        SetFit & \textbf{800} & 0.6808 & 0.6709 & \textbf{0.6781} \\ \hline
        Vanilla & 14379 & 0.7083 & 0.6632 & 0.6780 \\ 
        SetFit & \textbf{1200} & 0.7104 & 0.6962 & \textbf{0.7054} \\ \hline
        Vanilla & 18734 & 0.7190 & 0.6712 & 0.6864 \\
        SetFit & \textbf{1600} & 0.7206 & 0.7097 & \textbf{0.7183} \\ \hline
    \bottomrule
    \end{tabular}
\end{table}

As can be seen from the table above, SetFit's contrastive learning approach yielded a better F1-score compared to the vanilla finetuning, despite only using a fraction of the training samples. 

Additionally, we observed that Weighted-F1 displays a larger gap between models compared to Micro-F1. This is particularly expected, since the problem of unbalanced data is exacerbated in the vanilla finetuning setup as the maximum number of samples per label increases.

\begin{figure*}
  \centering
  \includegraphics[scale=0.16]{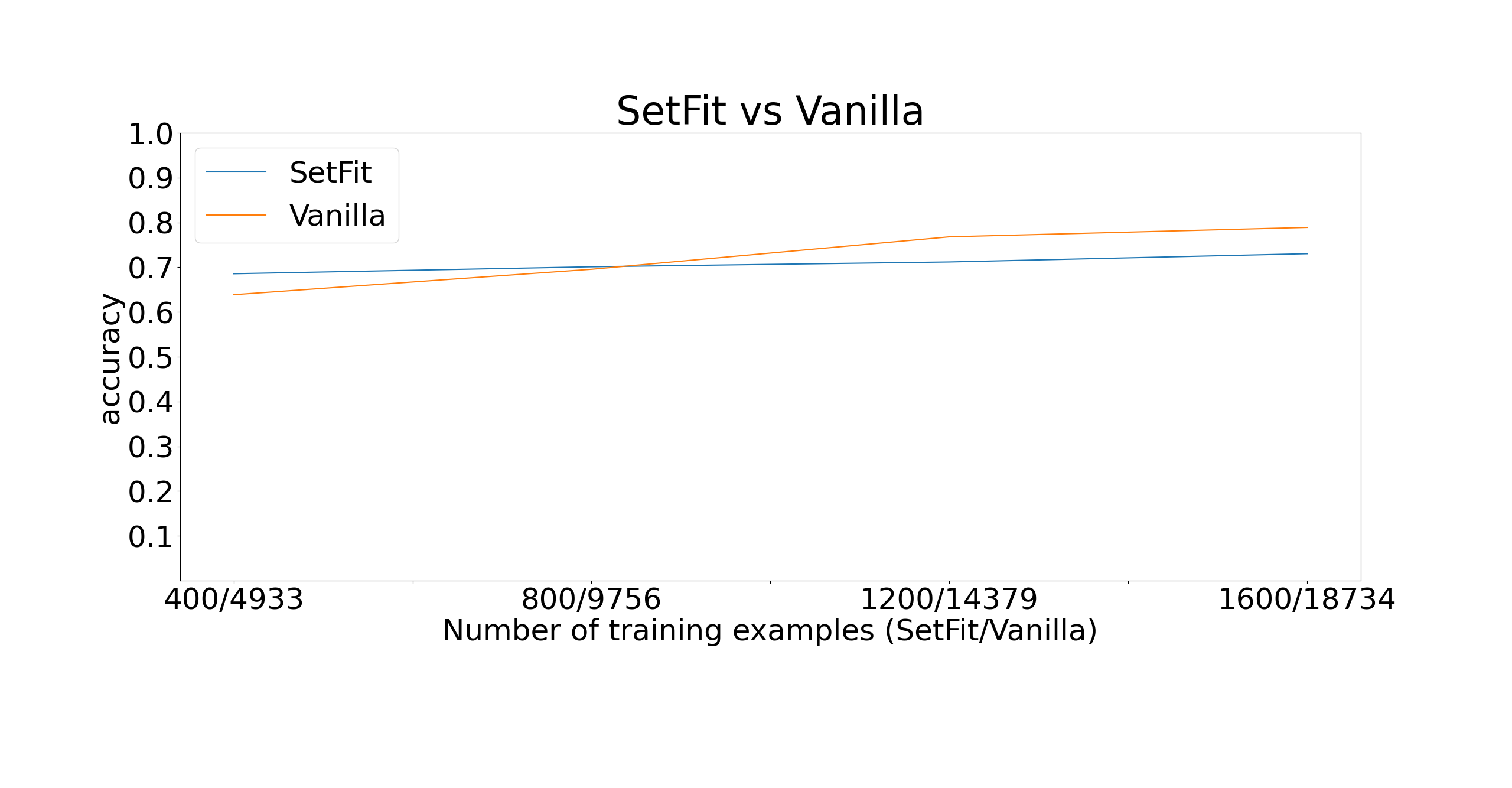}
  \caption{Accuracy comparison between SetFit and Vanilla finetuning, original LEDGAR dataset}
  \label{fig:acc_org}
\end{figure*}

\begin{figure*}
  \centering
  \includegraphics[scale=0.16]{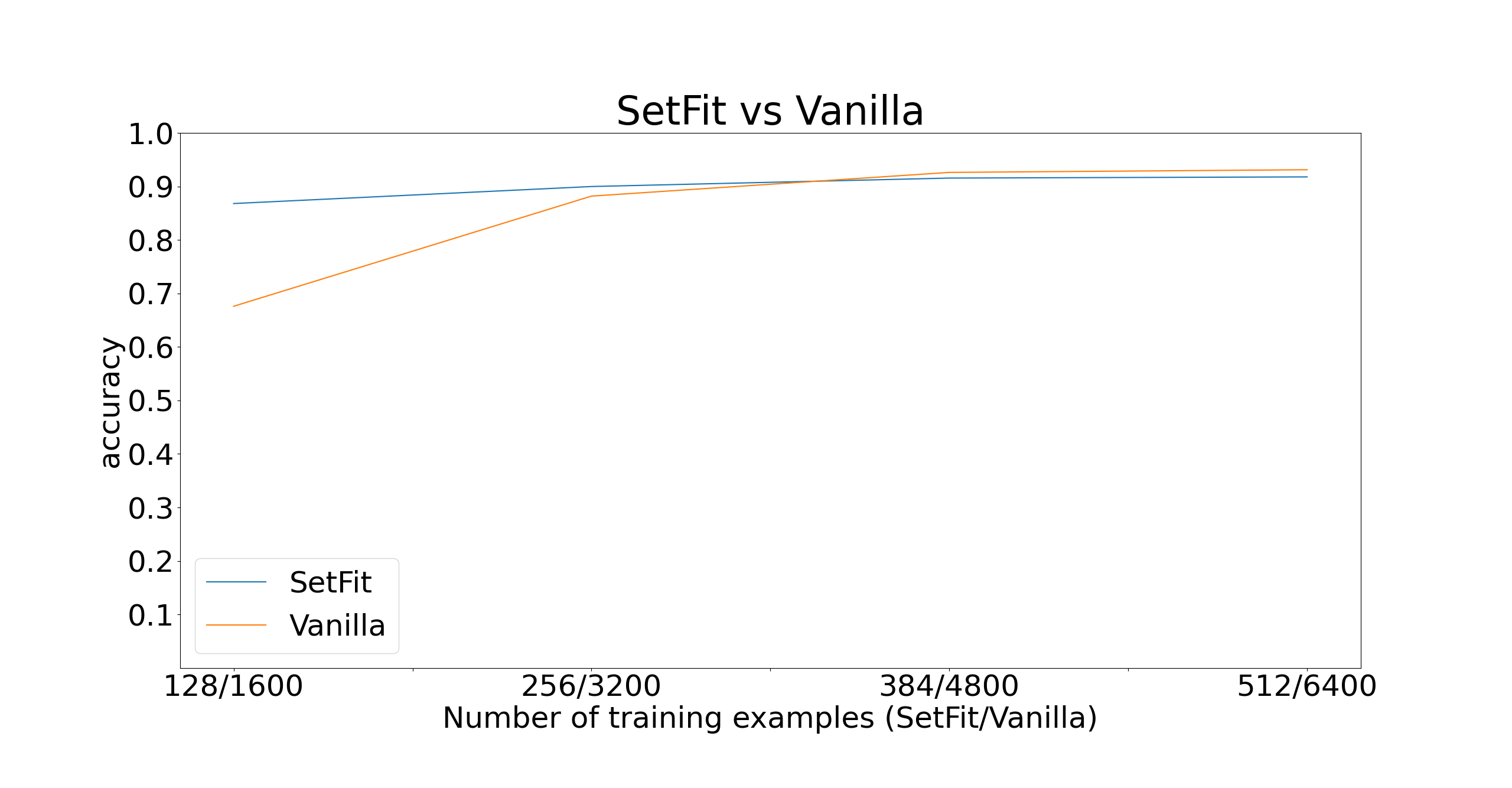}
  \caption{Accuracy comparison between SetFit and Vanilla finetuning, balanced LEDGAR dataset}
  \label{fig:acc_balanced}
\end{figure*}

\subsection{Accuracy comparisons: Original and balanced dataset}
In Figure~\ref{fig:acc_org}, we compare the finetuning results of SetFit and vanilla models, finetuned on the original LEDGAR dataset with the same training split and test dataset as the previous experiment. 

We observed that the models achieve comparable accuracies overall, despite the differences in Weighted F1-scores in the Table~\ref{tab:f1-score}. However, it is still noteworthy that the contrastive learning approach achieves accuracy comparable to the vanilla finetuned model with very small sample sizes.

In Figure~\ref{fig:acc_balanced}, we compare the accuracy of the two approaches, this time with the balanced LEDGAR dataset. In this experiment we also used 80/20 train/dev split. 
The results show that the contrastive learning finetuning has a warmer start compared to vanilla finetuning, particularly in small data scenarios. However, as can be seen from the graph, SetFit is comparable with vanilla model across all the experiments as well.

\subsection{LIME feature comparisons}
In machine learning in general, but especially in domains such as law, trustworthiness of AI systems is crucial. The ability to explain model predictions is central to increasing trustworthiness in at least two respects. First, explanations have an impact on whether a user can trust the prediction of the model to act upon it; second, they also influence whether a user can trust the model to behave in a certain way when deployed.

Several approaches to explaining model predictions have been proposed in the literature, including LIME~\cite{lime}, SHAP~\cite{DBLP:journals/corr/LundbergL17}, and GRAD-CAM~\cite{DBLP:journals/corr/SelvarajuDVCPB16}. 

Through the training results mentioned in previous sections, we observed that SetFit models were comparable with vanilla models, despite using a fraction of the dataset. However, we get very little information about whether the models base their decisions on features which are intuitively correct, that is, if the models are classifying the provisions with legally informative features, or arbitrary ones. 

LIME is a technique based on the creation of interpretable, surrogate models over the features that are locally faithful to the original classifier. This means that interpretable explanations need to use representations of those features that are understandable, trustworthy, and justifiable to humans~\cite{lime}.

For the text classification tasks, LIME features are restricted to the words that are presented in the provisions. Thus, the positively weighted words that lead toward a particular label are called "positive" features. Likewise, the negatively weighted words that reduce the model's estimate of the probability of the label are called "negative" features.

We kept the LIME hyperparameters the same in each model explanation for fair comparison and the details are as follows: The limit for the total number of words per classification is defined as $K$, and the complexity measure for the models is defined as:
\begin{displaymath}
\Omega(g) = \infty\mathbb{1}[\lVert w_g\rVert_0 > K]
\end{displaymath} where the $g$ is defined as a simple interpretable sparse linear model (logistic regression in the case of SetFit, multinomial logistic regression in the vanilla model); $w_g$ is defined as the weight vector of $g$. The $K=10$ is selected across all the experiments for simplicity and potentially can be as big as the computation allows. The size of the neighborhood for local exploration is set to 25. The distance function $D$ was kept as the default, cosine distance metric. \footnote{https://github.com/marcotcr/lime}

Thus, in this section, we compare the positive and negative features of SetFit and vanilla models extracted using the LIME setup mentioned above. 

To ensure a fair comparison, we used the SetFit model trained with 800 training samples and the vanilla model trained with 9756 training samples. As shown in Figure~\ref{fig:acc_org}, the two models converged and obtained comparable performance with these settings. We selected two test labels to compare, namely Adjustments and Authority provisions. Again, for a fair comparison, we chose the labels based on the cases where one technique did better than the other, in terms of their respective F1-scores. For the Adjustments label, the SetFit model outperformed the vanilla model, and for the Authority label, vanilla finetuning outperformed the SetFit model. Thus, we aim to observe the differences in the model-predicted features for these labels. Table~\ref{tab:lime-labels} shows the F1-score differences of these provisions.

\begin{table}
    \centering
    \caption{F1-score comparisons of Adjustments and Authority provisions}
    \label{tab:lime-labels}
    \begin{tabular}{ccc}
        \toprule
        Class Label & Vanilla & SetFit \\ \hline
        \midrule
        Adjustments & 0.7368 & \textbf{0.8571}\\ \hline
        Authority & \textbf{0.5063} & 0.2903 \\ 
    \bottomrule
    \end{tabular}
\end{table}

\begin{figure*}
  \centering
  \includegraphics[scale=0.21]{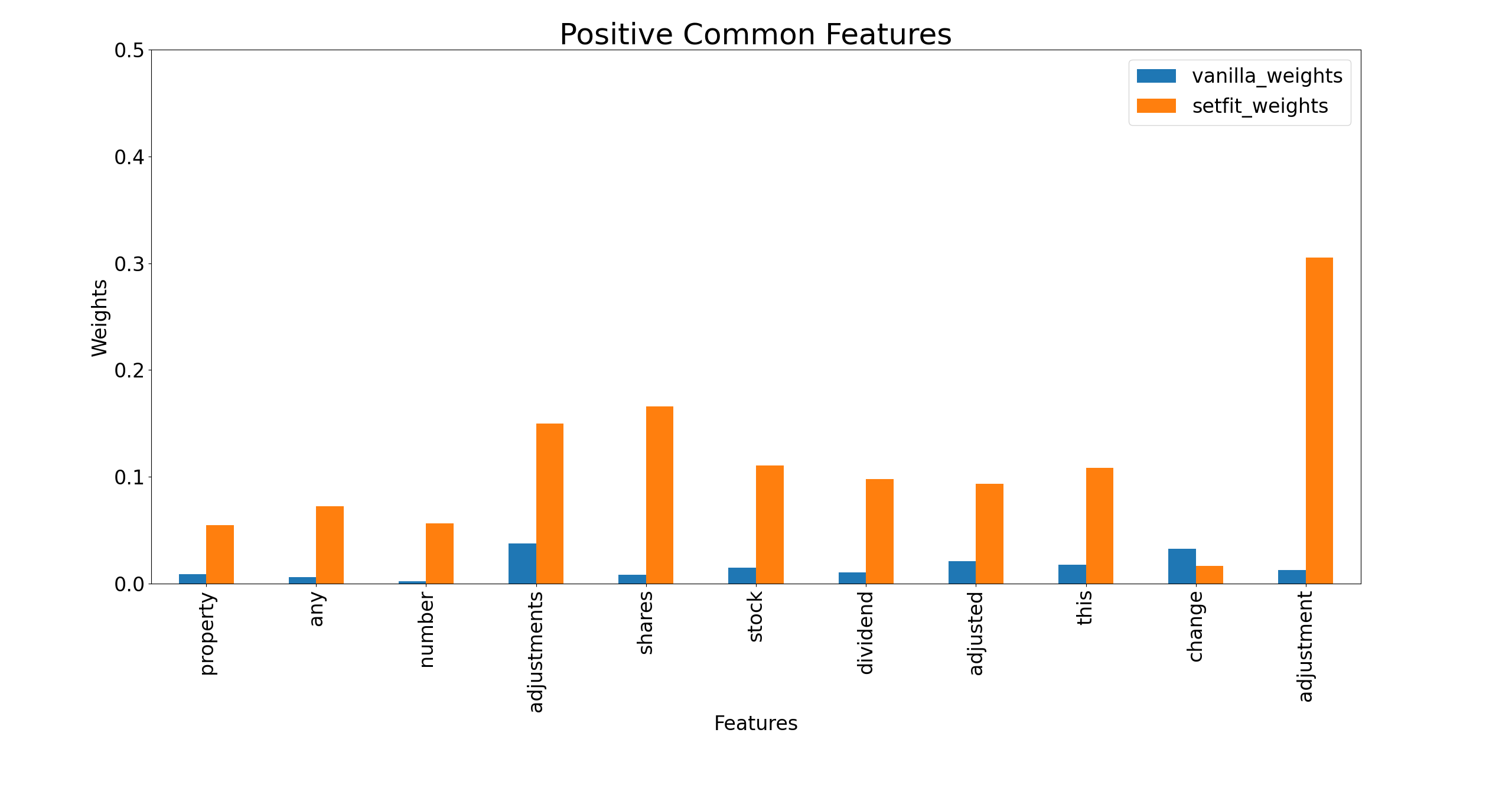}
  \caption{SetFit vs Vanilla finetuning, common positive LIME features comparison for Adjustments provision}
  \label{fig:positive_common_adjustments}
\end{figure*}

\begin{figure*}
  \centering
  \includegraphics[scale=0.21]{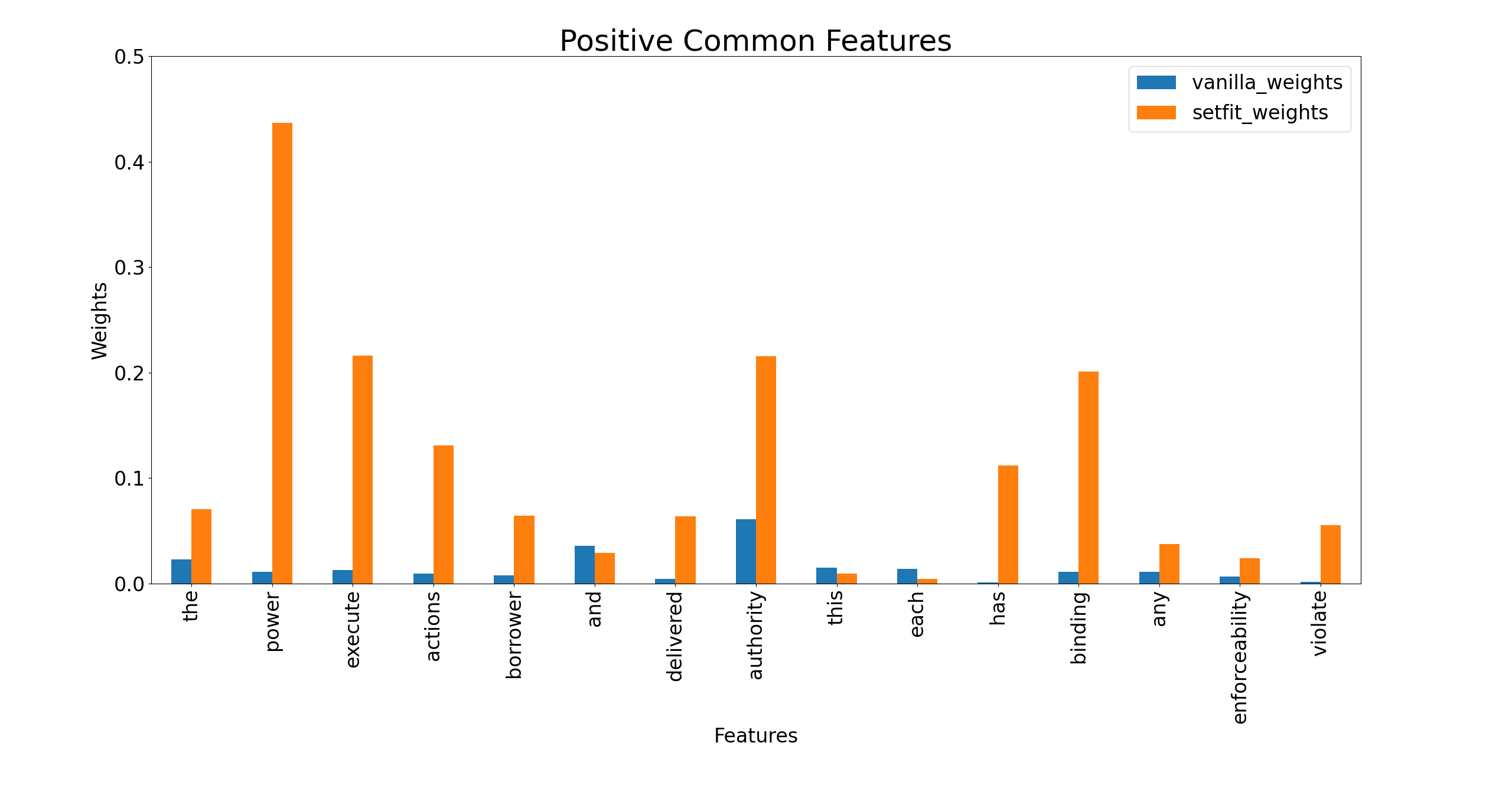}
  \caption{SetFit vs Vanilla finetuning, common positive LIME features comparison for Authority provision}
  \label{fig:positive_common_authority}
\end{figure*}

We begin by comparing the positive features which both approaches have in common (i.e. the features they both assign a positive weight to), for the two target labels. These are shown in Figure \ref{fig:positive_common_adjustments} and Figure \ref{fig:positive_common_authority}. The figures suggest that the contrastive approach from SetFit seems to help to boost legally informative features more than vanilla models, even in the small data scenarios. For instance, words like \textit{"adjustments"}, \textit{"shares"}, \textit{"dividend"}, \textit{"stock"}, etc. can give a first strong hint about the Adjustment provision classification results, as well as words like \textit{"authority"}, \textit{"power"}, \textit{"act"}, \textit{"execute"}, \textit{"binding"}, etc. for the Authority provision. Thus, domain experts can make decisions based on their usefulness. 

\begin{figure*}
  \centering
  \includegraphics[scale=0.21]{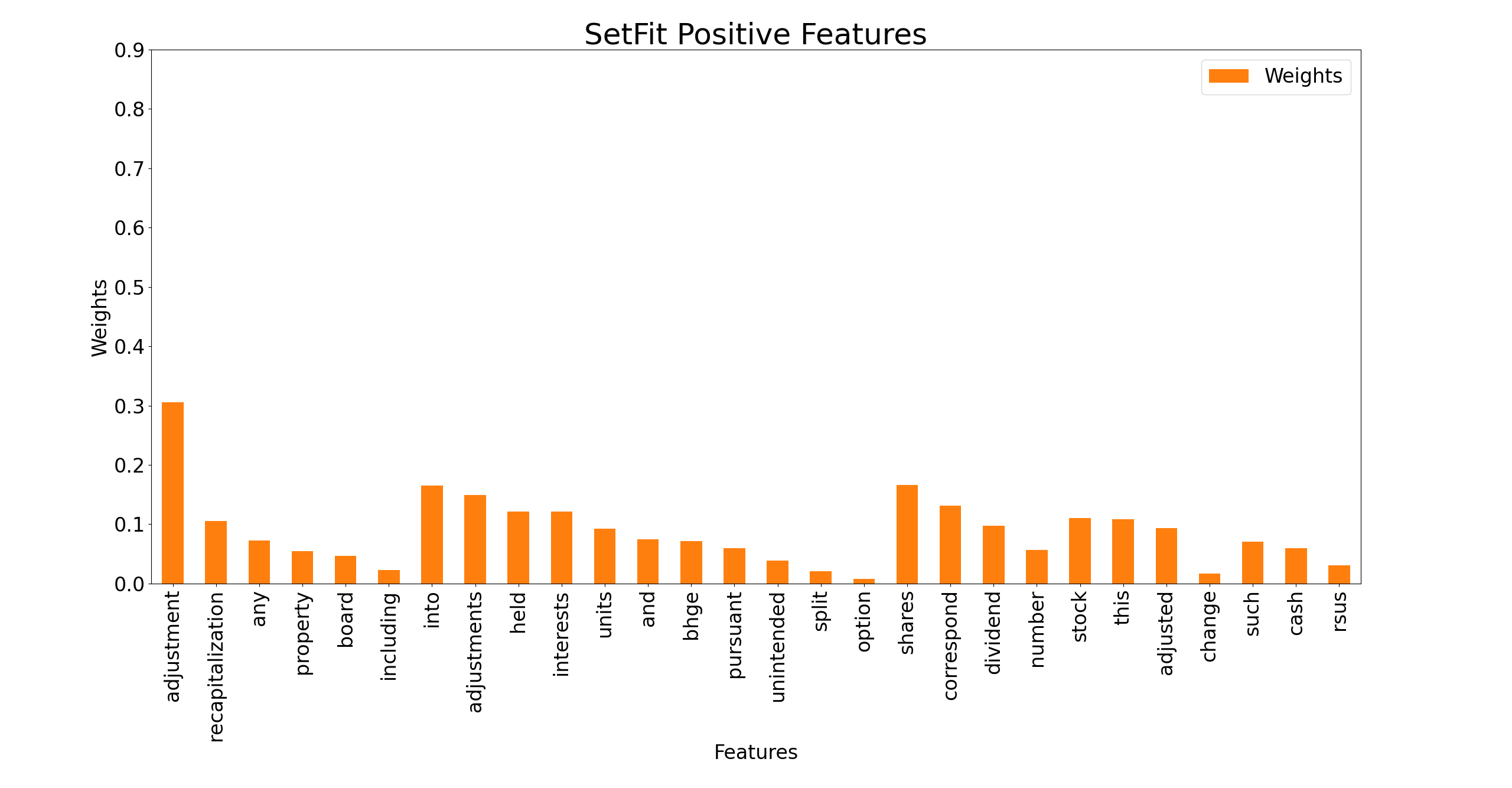}
  \caption{SetFit finetuning positive LIME features for Adjustments provision}
  \label{fig:positive_adjustments_setfit}
\end{figure*}

\begin{figure*}
  \centering
  \includegraphics[scale=0.21]{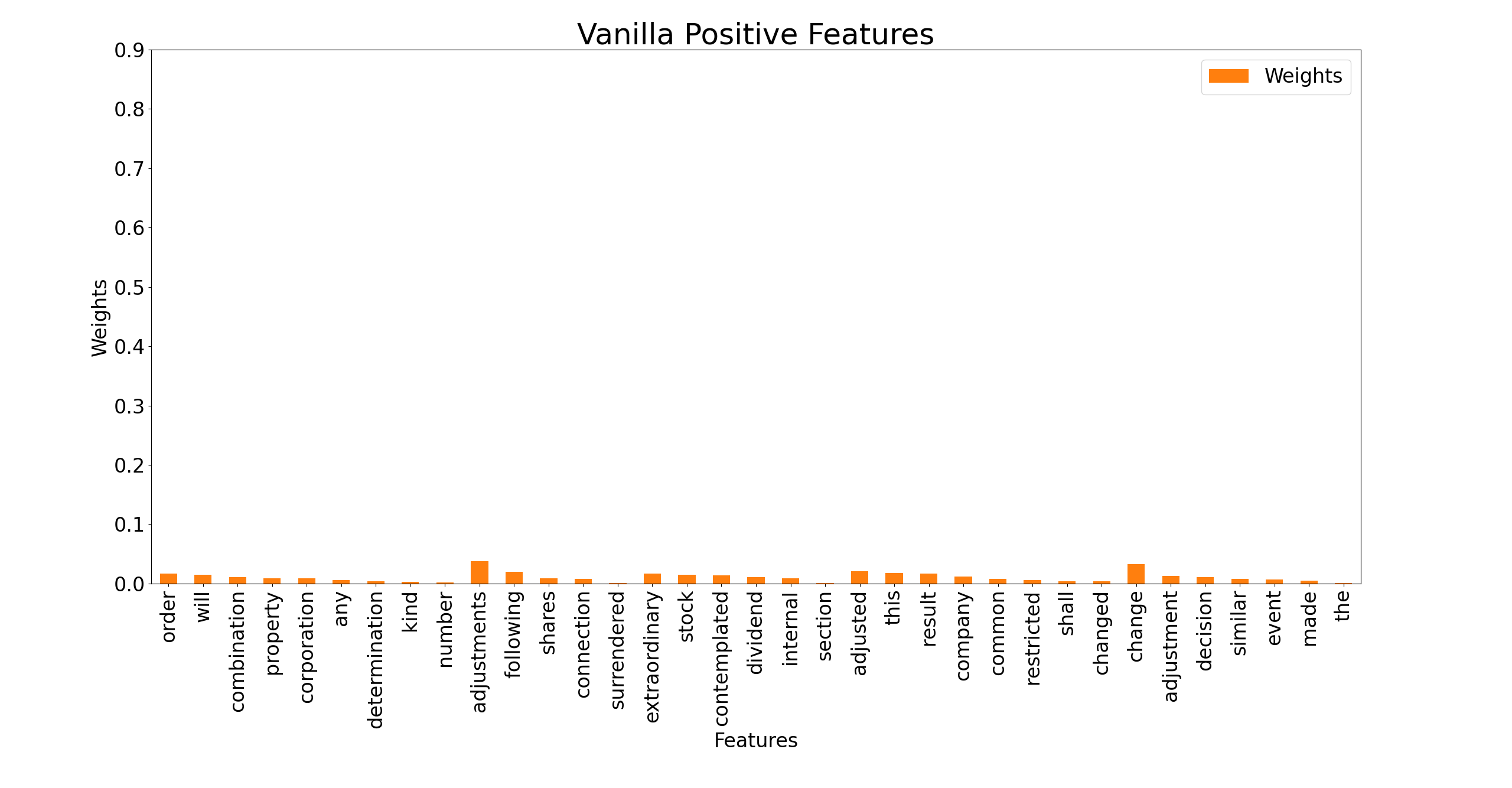}
  \caption{Vanilla finetuning positive LIME features for Adjustments provision}
  \label{fig:positive_adjustments_vanilla}
\end{figure*}

\begin{figure*}
  \centering
  \includegraphics[scale=0.21]{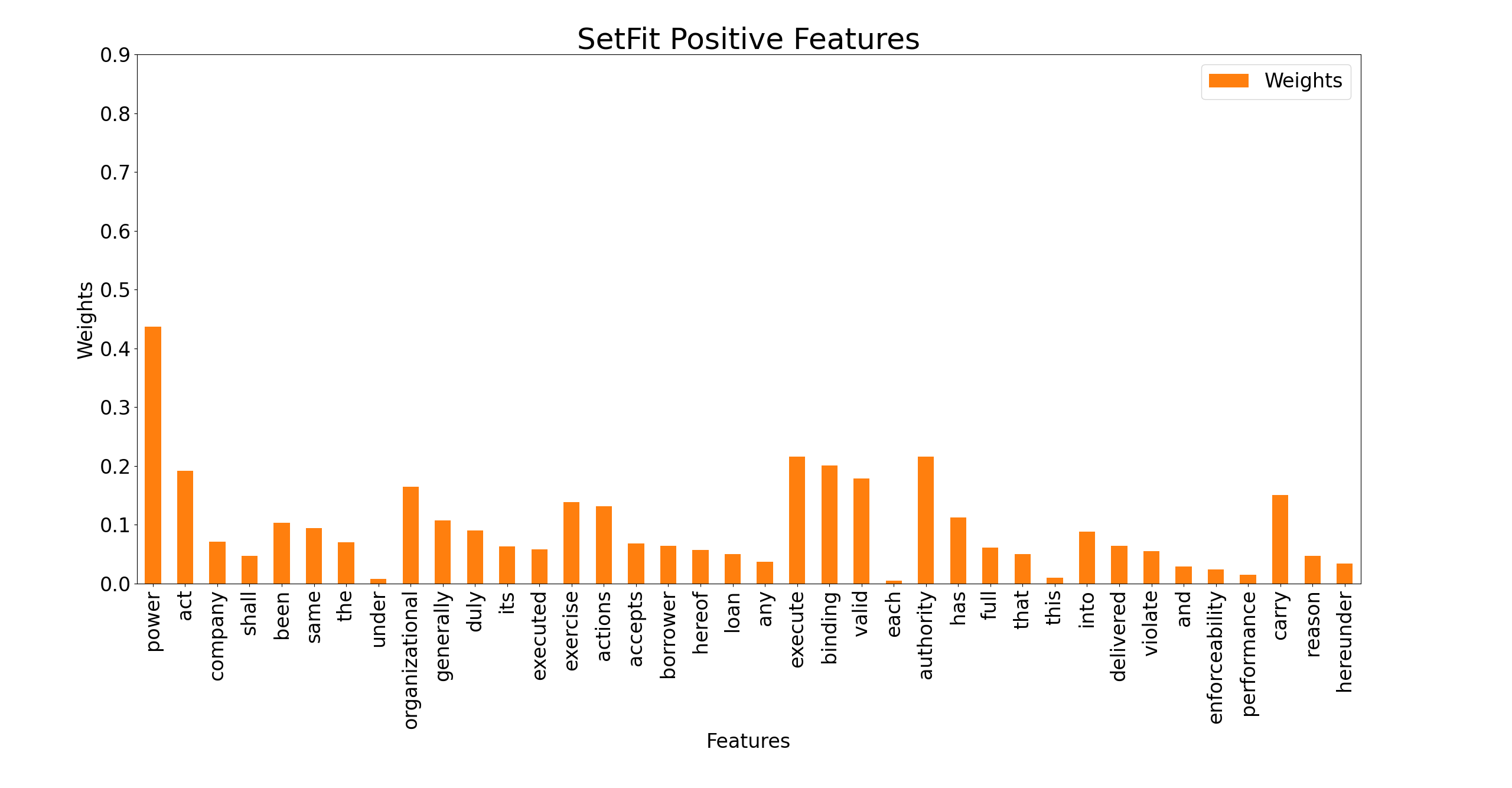}
  \caption{SetFit finetuning positive LIME features for Authority provision}
   \label{fig:positive_authority_setfit}
\end{figure*}

\begin{figure*}
  \centering
  \includegraphics[scale=0.21]{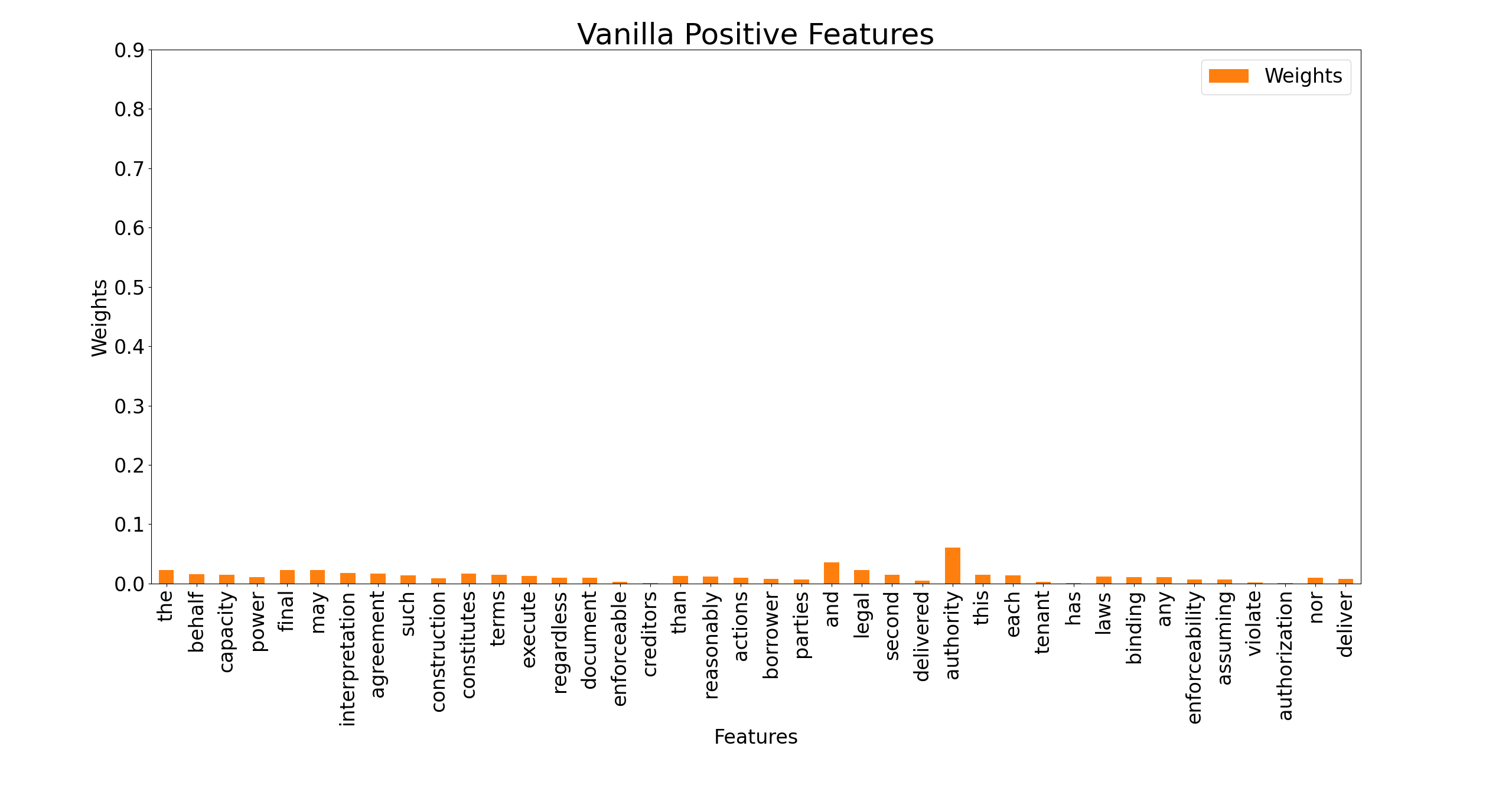}
  \caption{Vanilla finetuning positive LIME features for Authority provision}
   \label{fig:positive_authority_vanilla}
\end{figure*}

In Figures~\ref{fig:positive_adjustments_setfit} to~\ref{fig:positive_authority_vanilla}, we show the top positive features for the two models separately, for each label. We note that similar observations can be made with respect to these figures, that is, the contrastive learning framework boosts the positive weight of features that are intuitively more legally informative. Nevertheless, we also see that less informative features, including stop words, are also assigned some positive weight.

\begin{figure*}
  \centering
  \includegraphics[scale=0.21]{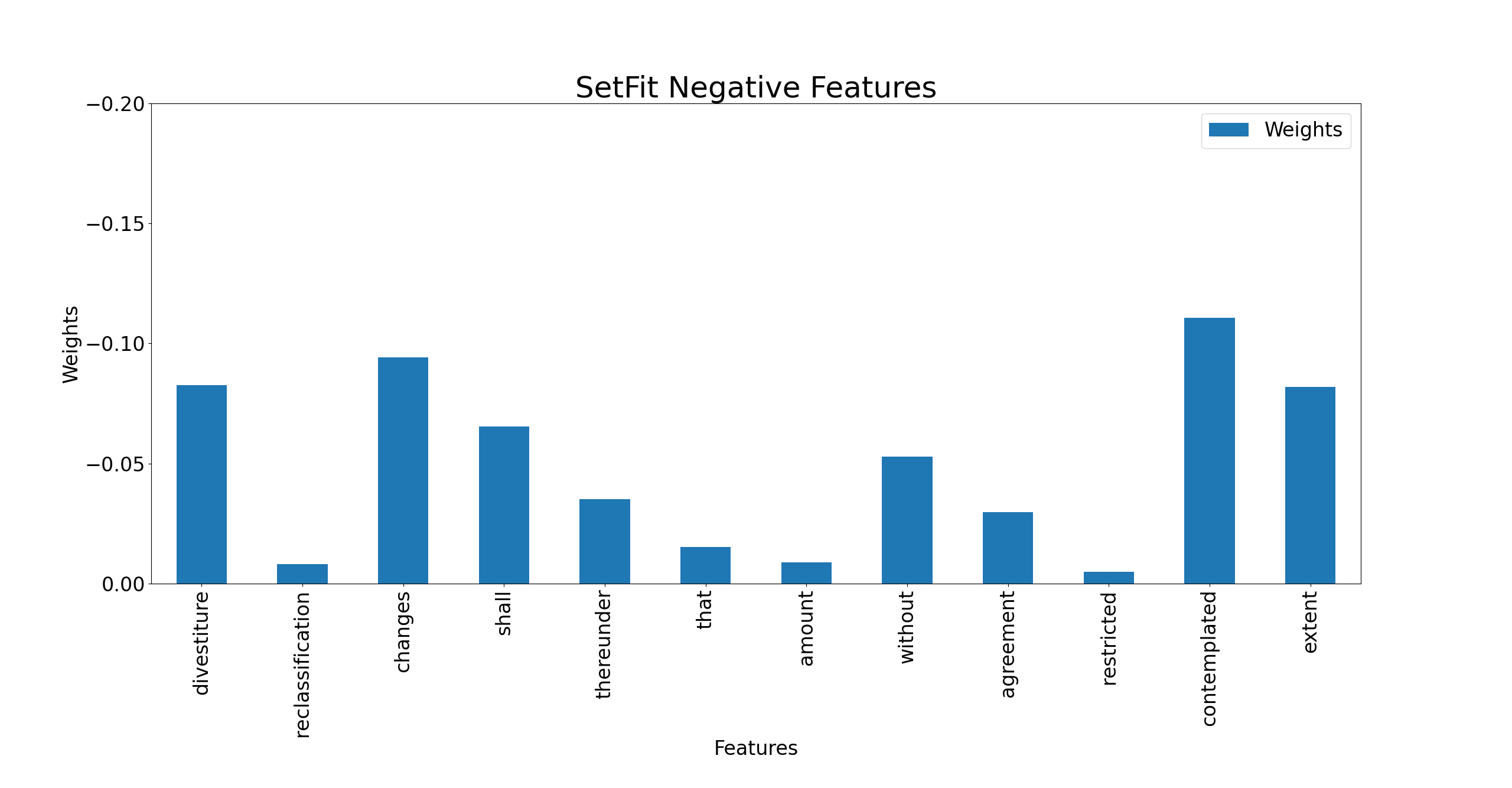}
  \caption{SetFit finetuning negative LIME features for Adjustments provision}
  \label{fig:negative_adjustments_setfit}
\end{figure*}

\begin{figure*}
  \centering
  \includegraphics[scale=0.21]{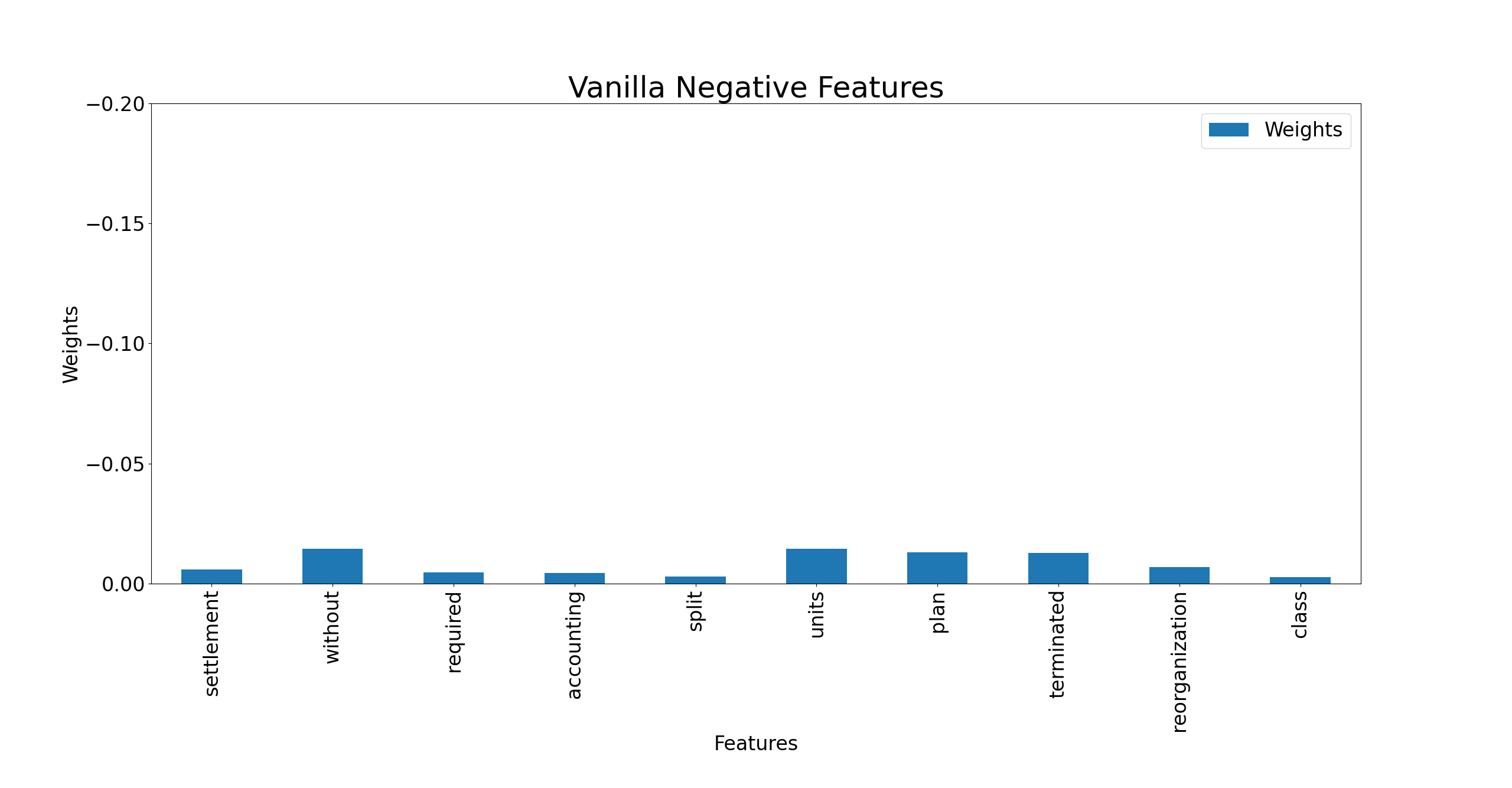}
  \caption{Vanilla finetuning, negative LIME features for Adjustments provision}
  \label{fig:negative_adjustments_vanilla}
\end{figure*}

\begin{figure*}
  \centering
  \includegraphics[scale=0.21]{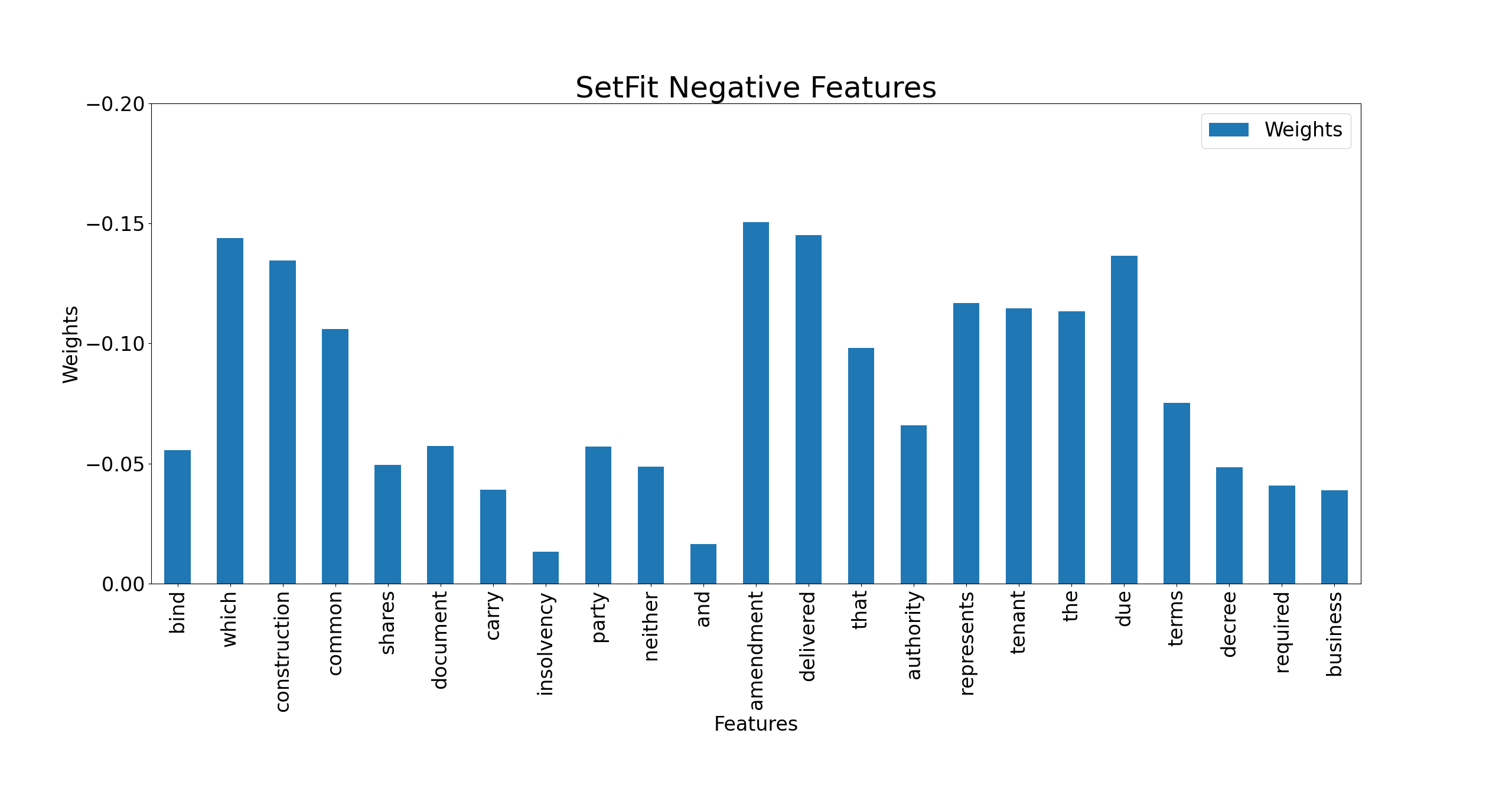}
  \caption{SetFit finetuning negative LIME features for Authority provision}
  \label{fig:negative_authority_setfit}
\end{figure*}

\begin{figure*}
  \centering
  \includegraphics[scale=0.21]{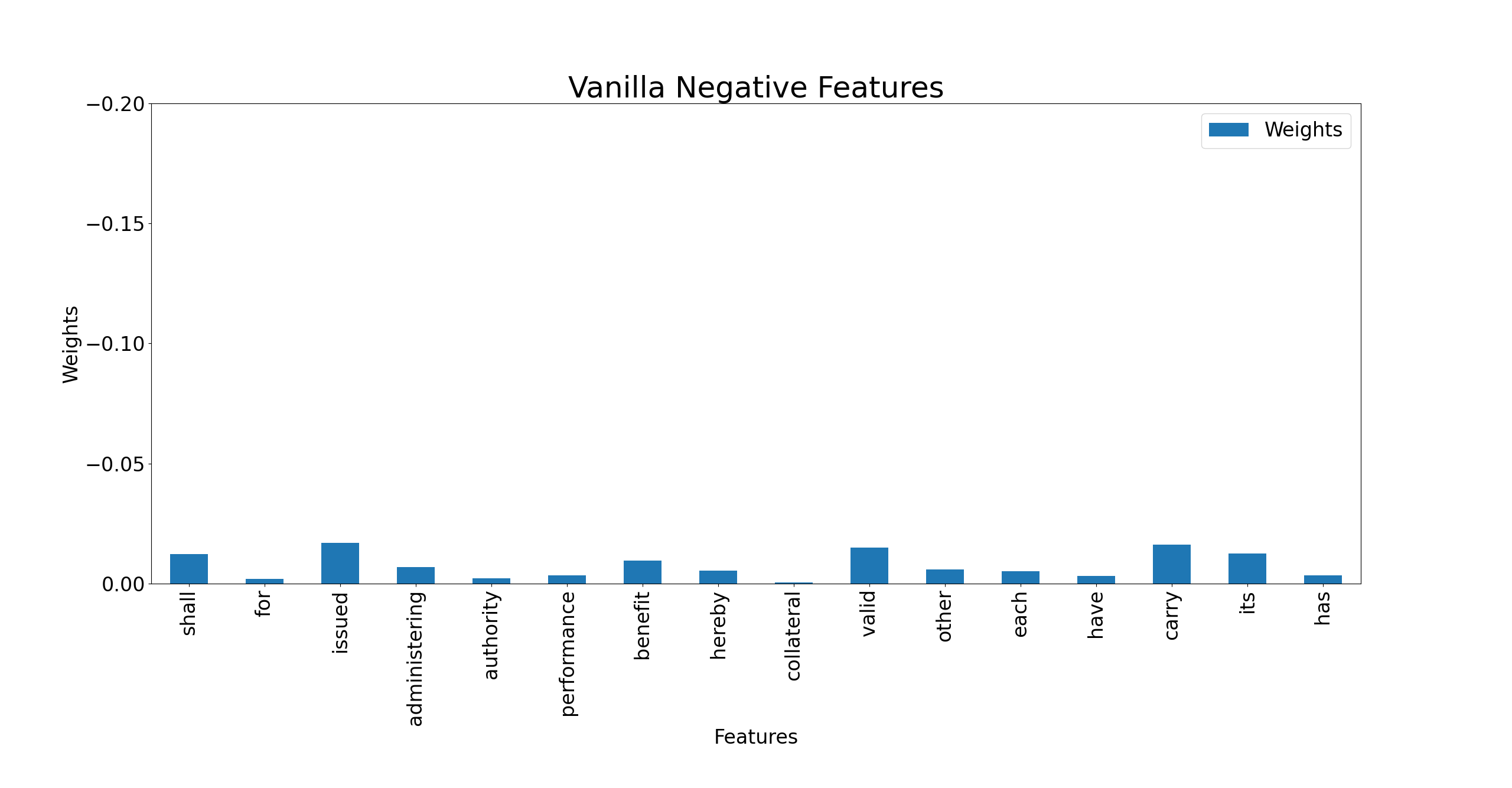}
  \caption{Vanilla finetuning, negative LIME features for Authority provision}
   \label{fig:negative_authority_vanilla}
\end{figure*}

Additionally, we also observed similar behavior with the negative features in Figures~\ref{fig:negative_adjustments_setfit} to~\ref{fig:negative_authority_vanilla}. For negative features, the SetFit model trained with a contrastive objective assigns a greater negative magnitude. Thus, it appears that negative role of these features is accentuated in the contrastive setting, relative to the standard finetuning setup. For instance, words like \textit{"changes"}, \textit{"shall"} and \textit{"without"} for the Adjustments provision and \textit{"which"}, \textit{"common"}, \textit{"document"} and \textit{"carry"} for the Authority provision sound generic and may not give legally informative hints to humans. However, in the vanilla model case, similar legally non-informative negative features are also present but not enough to perturb the model's decisions.

\section{Conclusions \& Future Work}

This paper presented a detailed comparison of legal provision classification. Motivated by the challenge of low-resource scenarios and data imbalance, we compared the performance of a LegalBERT model finetuned in a standard setting, to one finetuned using a contrastive objective. 

Following previous work~\cite{chalkidis-etal-2020-legal}, we assumed that models pretrained on legal data are better able to retain the legal knowledge and terminologies in the process of finetuning. On the other hand, our experiments show that the type of finetuning approach matters, especially where data is relatively scarce. In particular, the contrastive learning approach showed promising results in terms of evaluation metrics, achieving performance comparable or better than the vanilla finetuning setup. The results also showed that the positive and negative features extracted from the models differ significantly, favoring the SetFit model, despite using almost 11 times less data.

As future work, investigating the limitations of SetFit deeper with more hyperparameters on legal data may be beneficial for pushing the model capabilities further. Also, we plan to use other explainability tools such as SHAP or GRAD-CAM to compare the extracted features. Finally, an evaluation of the appropriateness of the positive and negative features identified using explainability methods needs to be carried out with domain experts.

\clearpage
\bibliography{contrast_is_all_you_need}

\end{document}